\relax
\documentclass[letterpaper]{article} %DO NOT CHANGE THIS
\usepackage{aaai18}  %Required
\usepackage{times}  %Required
\usepackage{helvet}  %Required
\usepackage{courier}  %Required
\usepackage{url}  %Required
\usepackage{graphicx}  %Required
\usepackage{subcaption}
\usepackage{paralist,color}

\usepackage[linesnumbered,ruled,vlined]{algorithm2e}
\usepackage{algpseudocode}

\begin{document}
% The file aaai.sty is the style file for AAAI Press 
% proceedings, working notes, and technical reports.
%
\newtheorem{prop}{Proposition}
\def\x{x}
\def\y{y}
\def\z{z}
\def\R{\RR}
\def\Z{\cZ}
\def\U{\cU}
\newcommand{\bx}{{\boldsymbol{x}}}

\newcommand{\alert}[1]{\textcolor{red}{\bf #1}}
\newcommand{\best}[1]{{\bf #1}}

\def\astro{{\sf astro-physic}\xspace}
\def\real{{\sf real-sim}\xspace}
\def\news{{\sf news20}\xspace}
\def\kddb{{\sf kddb}\xspace}
\def\e2006{{\sf TFIDF-2006}\xspace}
\def\webspam{{\sf webspam}\xspace}
\def\japan{{\sf yahoo-japan}\xspace}
\def\rcv1{{\sf rcv1}\xspace}
\def\korea{{\sf yahoo-korea}\xspace}
\def\a9a{{\sf a9a}\xspace}
\def\covtype{{\sf covtype}\xspace}
\def\ijcnn{{\sf ijcnn1}\xspace}

\def\sdcd{{\it DCD}\xspace}
\def\cocoa{{\it CoCoA}\xspace}
\def\ascd{{\it AsySCD}\xspace}
\def\asdcd{{\it PASSCoDe}\xspace} 
\def\atomic{{\it PASSCoDe-Atomic}\xspace} 
\def\wild{{\it PASSCoDe-Wild}\xspace} 
\def\lock{{\it PASSCoDe-Lock}\xspace} 
\def\bfatomic{{\bf PASSCoDe-Atomic}} 
\def\bfwild{{\bf PASSCoDe-Wild}} 
\def\bflock{{\bf PASSCoDe-Lock}} 

\newlength\myindent
\setlength\myindent{2em}
\newcommand\bindent{%
  \begingroup
  \setlength{\itemindent}{\myindent}
  \addtolength{\algorithmicindent}{\myindent}
}
\newcommand\eindent{\endgroup}

\title{ImageNet Training in Minutes}
\author{Yang You$^1$, Zhao Zhang$^2$, Cho-Jui Hsieh$^3$, James Demmel$^1$, Kurt Keutzer$^1$\\
UC Berkeley$^1$, TACC$^2$, UC Davis$^3$\\
\{youyang, demmel, keutzer\}@cs.berkeley.edu; zzhang@tacc.utexas.edu; chohsieh@ucdavis.edu \\
%Palo Alto, California 94303\\
}
\maketitle
\begin{abstract}
Since its creation, the ImageNet-1k benchmark set has played a significant role as a benchmark for ascertaining the accuracy of different deep neural net (DNN) models on the classification problem. Moreover, in recent years it has also served as the principal benchmark for assessing different approaches to DNN training. Finishing a 90-epoch ImageNet-1k training with ResNet-50 on a NVIDIA M40 GPU takes 14 days.
This training requires $10^{18}$ single precision operations in total.
On the other hand, the world's current fastest supercomputer can finish $2 \times 10^{17}$ single precision operations per second.
If we can make full use of the computing capability of the fastest supercomputer for DNN training, we should be able to finish the 90-epoch ResNet-50 training in five seconds.
Over the last two years, a number of researchers have focused on closing this significant performance gap through scaling DNN training to larger numbers of processors. 
Most successful approaches to scaling ImageNet training have used the synchronous stochastic gradient descent.
However, to scale synchronous stochastic gradient descent one must also increase the batch size used in each iteration.

Thus, for many researchers, the focus on scaling DNN training has translated into a focus on developing training algorithms that enable increasing the batch size in data-parallel synchronous stochastic gradient descent without losing accuracy over a fixed number of epochs. 
As a result, we have seen the  batch size and number of processors successfully utilized  increase from 1K batch size on 128 processors to 8K batch size on 256 processors over the last two years. 
The recently published LARS algorithm  
increased batch size further to 32K for some DNN models. 
Following up on this work, we wished to confirm that LARS could be used to further scale the number of processors efficiently used in DNN training and, and as a result, further reduce the total training time.
%In this paper we present the results of this investigation: using LARS we were able to efficiently utilize 512 KNL chips to finish the 100-epoch ImageNet training with AlexNet in 24 minutes, and
In this paper we present the results of this investigation: using LARS we efficiently utilized 1024 CPUs to finish the {\bf 100-epoch} ImageNet training with {\bf AlexNet in 11 minutes with 58.6\% accuracy (batch size = 32K)}, and
we utilized 2048 KNLs to finish the 90-epoch ImageNet training with ResNet-50 in 20 minutes without losing accuracy (batch size = 32K).
State-of-the-art ImageNet training speed with ResNet-50 is 74.9\% top-1 test accuracy in 15 minutes \cite{akiba2017extremely}.
{\bf We got 74.9\% top-1 test accuracy in 64 epochs, which only needs 14 minutes}.
%Both of these training runs use a batch size of 32K.
%However, our hardware budget is only 1.2 million USD, which is 3.4 times lower than Facebook's 4.1 million USD.
Furthermore, when the batch size is above 16K, our accuracy using LARS is much higher than Facebookâs corresponding batch sizes (Figure \ref{fig:facebook_compare}).
Our code is available upon request. %It will also be released in Intel Caffe.
\end{abstract}

\begin{figure}
\includegraphics[width=3.1in]{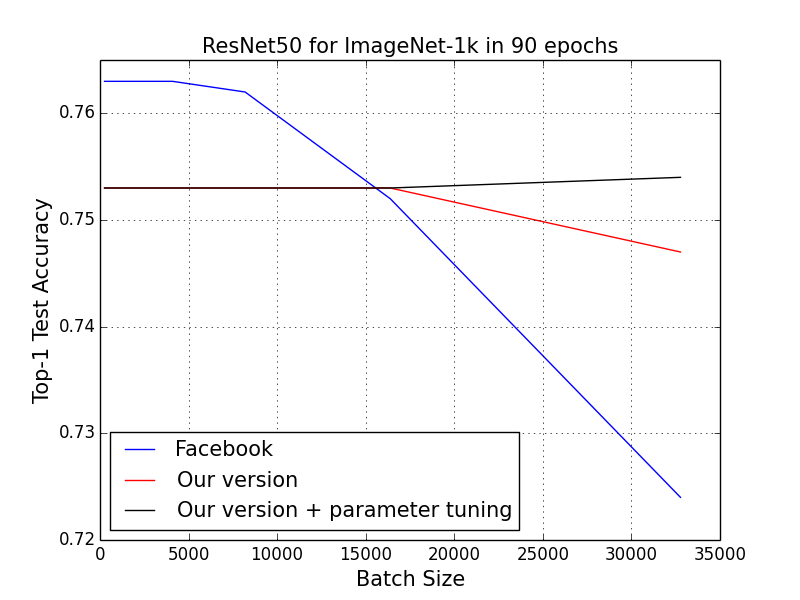}
\caption{\footnotesize \label{fig:facebook_compare}Because we use weaker data augmentation, our baseline's accuracy is slightly lower than Facebook's version (76.2\% vs 75.3\%). However, at very large batch sizes our accuracy is much higher than Facebook's accuracy. Facebook's accuracy is from their own report \cite{goyal2017accurate}. Our accuracy scaling efficiency is much higher than Facebook's version}
\end{figure}

\begin{table}[tb]
\footnotesize
  %\vspace{-10pt}
  \caption{Compare to state-of-the-art ImageNet training speed with ResNet-50.}
  \label{tab:compare_time}
\centering
    \vspace{3pt}
  \begin{tabular}{|c|ccc|}
    \hline
    Work & Batch Size &  Test Accuracy & Time\\
    \hline
    Akiba et al. & 32K &  74.9\% & 15 mins\\
    \hline
    Our version & 32K &  74.9\%  & 14 mins\\
    \hline
     \end{tabular}
%  \vspace{-15pt}
\end{table}

\section{Introduction}
\label{sec:intro}
For deep learning applications, larger datasets and bigger models lead to significant improvements in accuracy \cite{amodei2015deep}, but at the cost of longer training times.
Moreover, many applications such as computational finance, autonomous driving, oil and gas exploration, and medical imaging, will almost certainly require training data-sets with billions of training elements and terabytes of data. This highly motivates the problem of accelerating the training time of Deep Neural Nets (DNN). 
For example, finishing 90-epoch ImageNet-1k training with ResNet-50 on a NVIDIA M40 GPU takes 14 days.
This training requires $10^{18}$ single precision operations in total.
On the other hand, the world's current fastest supercomputer can finish $2 \times 10^{17}$ single precision operations per second \cite{dongarra2017}.
Thus, if we can make full use of the computing capability of a supercomputer for DNN training, we should be able to finish the 90-epoch ResNet-50 training in five seconds.
%However, the current bottleneck for fast DNN %training is in the details of the optimization %algorithm. Specifically, current batch sizes %(e.g. 512) are unable to make full use of large %clusters of processors 
So  far, the best results on scaling ImageNet training have used synchronous stochastic gradient descent (syncronous SGD). The synchronous SGD algorithm has many inherent advantages, but at the root of these advantages is \textit{sequential consistency}. Sequential consistency implies that all valid parallel implementations of the algorithm match the behavior of the sequential version. This property is invaluable during DNN design and during the debugging of optimization algorithms. 
Continuing to scale the synchronous SGD model to more processors requires ensuring that there is sufficient useful work for each processor to do during each iteration. This, in turn, requires increasing the batch size used in each iteration. 
For example engaging 512 processors in synchronous SGD on  a batch size of 1K would mean that each processor only processed a local batch of 2 images. If the batch size can batch size can be scaled to 32K then each processor processes a local batch of 64, and the computation to communication ratio can be more balanced.  

As a result, over the last two years we have seen a focus on increasing the batch size and number of processors used in the DNN training for ImageNet-1K, with a resulting reduction in training time. 
In the following discussion we briefly review relevant work where all details of batch size, processors, DNN model, runtime, and training set are defined in the publications. All of the following refer to training on ImageNet.

FireCaffe \cite{arxiv-firecaffe} \cite{iandola2016firecaffe} demonstrated scaling the training of GoogleNet to 128 Nvidia K20 GPUs with a batch size of 1K for 72 epochs and a total training time of 10.5 hours. 
Although large batch size can lead to a significant loss in accuracy, using a warm-up scheme coupled with a linear scaling rule, researchers at Facebook \cite{goyal2017accurate} were able to scale the training of ResNet 50 to 256 Nvidia P100's with a batch size of 8K and a total training time of one hour. 
Using a more sophisticated approach to adapting the learning rate in a method they named  the Layer-wise Adaptive Rate Scaling (LARS) algorithm \cite{you2017scaling}, researchers were able to scale the batch size to very large sizes, such as 32K, although only 8 Nvidia P100 GPUs were employed. A 3.4\% reduction in accuracy was attributed to the absence of data augmentation. 

Given the large batch sizes that the LARS algorithm enables, it was natural to ask: how much further can we scale the training of DNNs on ImagNet? 
This is the investigation that led to this paper. In particular, we found that using LARS we could scale DNN training on ImageNet to 1024 CPUs and finish the {\bf 100-epoch training with AlexNet in 11 minutes with 58.6\% accuracy}.
Furthermore, we could scale to 2048 KNLs and finish the {\bf 90-epoch ImageNet training with ResNet50 in 20 minutes without losing accuracy}. 
State-of-the-art ImageNet training speed with ResNet-50 is 74.9\% top-1 test accuracy in 15 minutes \cite{akiba2017extremely}.
{\bf We got 74.9\% top-1 test accuracy in 64 epochs, which only needs 14 minutes}.

{\bf Notes.}
This paper is focused on training large-scale deep neural networks on $P$ machines/processors. 
We use $w$ to denote the parameters (weights of the networks), $w^j$ to denote the local parameters on $j$-th worker, 
$\tilde{w}$ to denote the global parameter. When there is no confusion we use $\nabla w^j$ to denote the stochastic 
gradient evaluated at the $j$-th worker. 
All the accuracy means top-1 test accuracy. There is no data augmentation in all the results.

\section{Background and Related Work}
\label{sec:related}

\subsection{Data-Parallelism SGD}

\begin{figure}[htb!]
    %\centering
    \begin{subfigure}[b]{0.55\textwidth}
        \centering
        \includegraphics[height=1.3in]{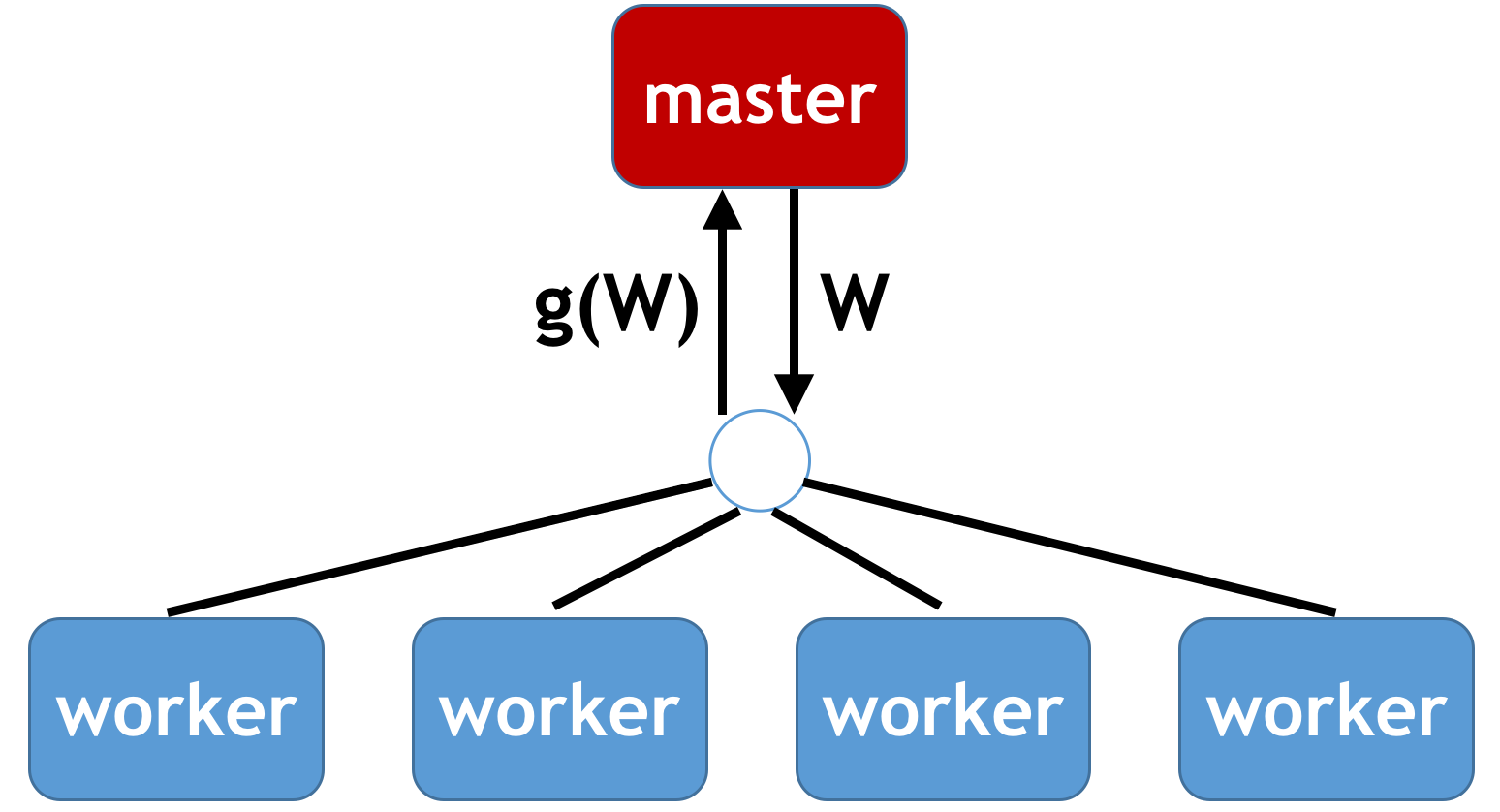}
        \caption{Data Parallelism}
    \end{subfigure}
    \begin{subfigure}[b]{0.55\textwidth}
        \centering
        \includegraphics[height=2.0in]{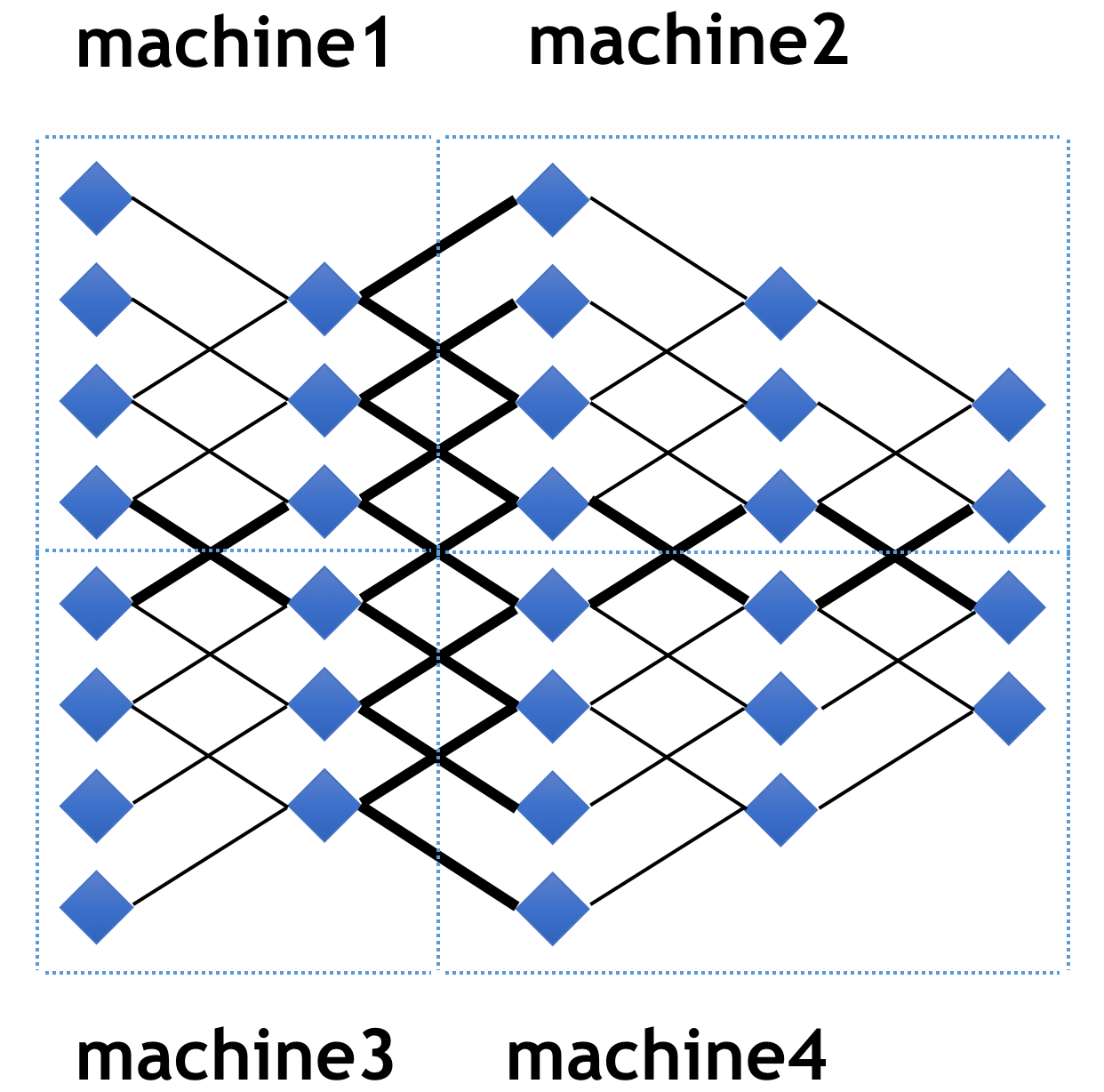}
        \caption{Model Parallelism}
    \end{subfigure}
    \caption{(a) is an example of data parallelism. Each worker sends its gradients $\nabla w^j$ to the master, and the master updates its weights by $\tilde{w} \leftarrow \tilde{w} -  \eta / P \sum_{i=1}^{P} \nabla w^j$. 
Then the master sends the updated weights $\tilde{w}$ to all the workers.
(b) is an example of model parallelism. A five layer neural network with
local connectivity is shown here, partitioned across four machines (blue rectangles). Only those
nodes with edges that cross partition boundaries (thick lines) will need to have their state communicated
between machines (e.g. by MPI \cite{gropp1996high}). Even in cases where a node has multiple edges crossing a partition boundary,
its state is only sent to the machine on the other side of that boundary once. %Within each partition, computation for individual nodes will the parallelized across all available processors (e.g. by OpenMP \cite{dagum1998openmp})
}
\label{fig:dl_parallelism}
\end{figure}

%There are two major directions for parallelizing DNN: data parallelism and model parallelism. 
%All the later parallel methods are the variants of them. 
%\subsubsection{Data Parallelism}
%{\bf Data Parallelism.}
In data parallelism method, the dataset is partitioned into $P$ parts stored on each machine, and each machine will have a local copy of the neural network and the weights ($w^j$). 
In synchronized data parallelism, the communication includes two parts: sum of local gradients and broadcast of the global weight. 
For the first part, each worker computes the local gradient $\nabla w^j$ independently, and 
sends the update to the master node. The master then updates $\tilde{w} \leftarrow \tilde{w} - \eta / P \sum_{j=1}^P \nabla w^j$ after it
gets all the gradients from workers. 
For the second part, the master broadcasts $\tilde{w}$ to all workers. 
This synchronized approach is a widely-used method on large-scale 
systems \cite{iandola2016firecaffe}.
Figure \ref{fig:dl_parallelism}-(a) is an example of 4 worker machines and 1 master machine.

Scaling synchronous SGD to more processors has two challenges. The first is giving each processor enough useful work to do; this has already been discussed. The second challenge is the inherent problem that after processing each local batch all processors must synchronize their gradient updates via a barrier before proceeding. This problem can be partially ameliorated by overlapping communication and communication \cite{das2016distributed} \cite{goyal2017accurate}, but the inherent synchronization barrier remains. 
A more radical approach to breaking this synchronization barrier is to pursue a purely asynchronous approach. 
A variety of asynchronous approaches have been proposed
\cite{recht2011hogwild} \cite{elastic-zhang2015deep} \cite{gossiping-jin2016scale} \cite{momentum-mitliagkas2016asynchrony}. 
The communication and updating rules differ in the asynchronous approach and the synchronous approach.
The simplest version of the asynchronous approach is a master-worker scheme. 
At each step, the master only communicates with one worker.
The master gets the gradients $\nabla w^j$ from the $j$-th worker, updates the global weights, and sends the global weight back to the $j$-th worker.
The order of workers is based on first-come-first-serve strategy. The master machine is also called as \textit{parameter server}.
The idea of a parameter server was used in real-world commercial applications by the Downpour SGD approach \cite{dean2012large}, which has 
succesfully scaled to $16,000$ cores. 
However, Downpour's performance on 1,600 cores for a globally connected network is not significantly better than a single GPU \cite{seide2014parallelizability}.

{\bf Model Parallelism}
Data parallelism replicates the neural network itself on each machine while model parallelism partitions the neural network into $P$ pieces. Partitioning the neural network means parallelizing the matrix operations on the partitioned network. Thus, model parallelism can get the same solution as the single-machine case. 
Figure \ref{fig:dl_parallelism}-(b) shows an example of using 4 machines to parallelize a 5-layer DNN. 
Model paralleism has been studied in~\cite{catanzaro2013deep,le2013building}. 
However, since the input size (e.g. size of an image) is relatively small, the matrix operations are not large. For example, parallelizing a 2048$\times$1024$\times$1024 matrix multiplication only needs one or two machines.  
Thus, state-of-the-art methods often use data-parallelism \cite{amodei2015deep,chen2016revisiting,dean2012large,seide20141}.

\section{Intel Knights Landing System}
\label{sec:knl}
Intel Knights Landing (KNL) is the latest version of Intel's general-purpose accelerator. The major distinct features of KNL that can benefit deep learning applications include the following:
{\bf (1) Self-hosted Platform.} The traditional accelerators (e.g. FPGA, GPUs, and KNC) rely on CPU for control and I/O management. 
%For some machine learning applications, the transfer path like PCIE may become a bottleneck at runtime because the memory on accelerator is limited (e.g. 12 GB GDDR5 on Nvidia K80 GPU). 
KNL does not need a CPU host. It is self-hosted by an operating system like CentOS 7.
{\bf (2) Better Memory.} KNL's measured bandwidth is much higher than that of a 24-core Haswell CPU (450 GB$/$s vs 100 GB$/$s). KNL's 384 GB maximum memory size is large enough to handle a typical deep learning dataset. Moreover, KNL is equipped with Multi-Channel DRAM (MCDRAM). MCDRAM's measured bandwidth is 475 GB/s. MCDRAM has three modes: a) Cache Mode: KNL uses it as the last level cache; b) Flat Mode: KNL treats it as the regular DDR; c) Hybrid Mode: part of it is used as cache, the other is used as the regular DDR memory.
{\bf (3) Configurable NUMA.}
The basic idea is that users can partition the on-chip processors and cache into different groups for better memory efficiency and less communication overhead. This is very important for complicated memory-access applications like DNN training.
%KNL supports all-to-all (A2A), quadrant/hemisphere (Quad/Hemi) and sub-NUMA (SNC-4/2) clustering modes of cache operation. For A2A, memory addresses are uniformly distributed across all tag directories (TDs) on the chip. For Quad/Hemi, the tiles are divided into four parts called quadrants, which are spatially local to four groups of memory controllers. Memory addresses served by a memory controller in a quadrant are guaranteed to be mapped only to TDs contained in that quadrant. Hemisphere mode functions the same way, except that the die is divided into two hemispheres instead of four quadrants. The SNC-4/2 mode partitions the chip into four quadrants or two hemispheres, and, in addition, expose these quadrants (hemispheres) as NUMA nodes. In this mode, NUMA-aware software can pin software threads to the same quadrant (hemisphere) that contains the TD and access NUMA-local memory.
%The architecture of KNL is in Figure \ref{fig:knl}.
%The architecture of KNL is in the appendix of this paper.

Since its release, KNL has been used in some HPC (High Performance Computing) data centers. For example, National Energy Research Scientific Computing Center (NERSC) has a supercomputer with 9,668 KNLs (Cori Phase 2). Texas Advanced Computing Center (TACC) has a supercomputer with 4,200 KNLs (Stampede 2).

In this paper, we have two chip options: (1) Intel Skylake CPU or (2) Intel KNL.
Using 1024 CPUs, we finish the 100-epoch AlexNet in 11 minutes and 90-epoch ResNet-50 in 48 minutes.
Using 1600 CPUs, we finish 90-epoch ResNet-50 in 31 minutes.
Using 512 KNLs, we finish the 100-epoch AlexNet in 24 minutes and 90-epoch ResNet-50 in 60 minutes.
Using 2048 KNLs, we finish 90-epoch ResNet-50 in 20 minutes.

\begin{figure}
\includegraphics[width=3.4in]{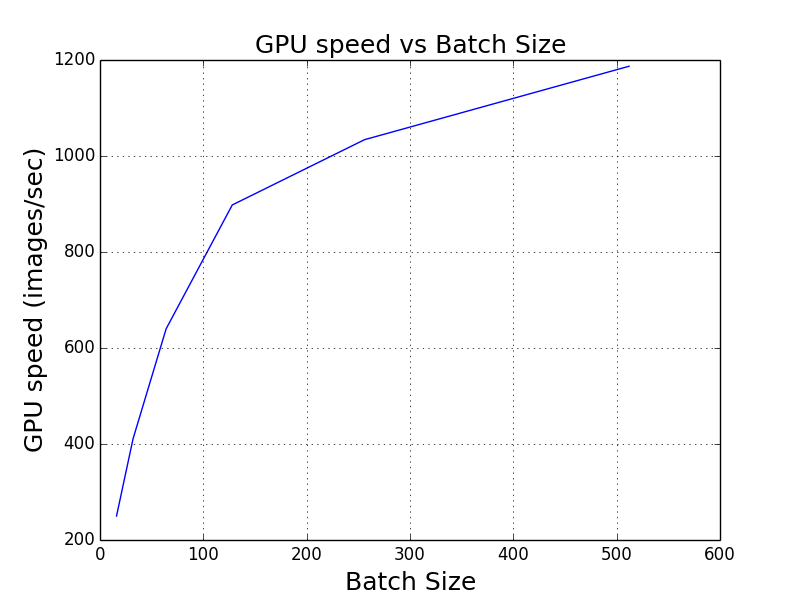}
\caption{\footnotesize \label{fig:gpu_speed}In a certain range, large batch improves the performance of system (e.g. GPU). The data in this figure is collected from training AlexNet by ImageNet dataset on NVIDIA M40 GPUs. Batch=512 per GPU gives us the highest speed. Batch=1024 per GPU is out of memory.}
\end{figure}

\begin{table*}[ht]
\footnotesize
%\tiny
\renewcommand{\arraystretch}{1.3}
\caption{Train neural networks by ImageNet dataset. $t_{comp}$ is the computation time and $t_{comm}$ is communication time. We fix the number of epochs as 100. Larger batch size needs much less iterations. Let us set batch size=512 per machine. Then we increase the number of machines. Since $t_{comp} \gg t_{comm}$ for using ImageNet dataset to train ResNet-50 networks and GPUs \cite{goyal2017accurate}, the single iteration time can be close to constant. Thus total time will be much less.}
\centering
\begin{tabular}{|c|ccccc|}
\hline
Batch Size & Epochs & Iterations & GPUs & Iteration Time & Total Time\\
\hline
\hline
512 & 100 & 250,000 & 1 & $t_{comp}$ & $250,000 \times t_{comp}$\\
\hline
1024 & 100 & 125,000 & 2 & $t_{comp}$ + log(2)$t_{comm}$ & $125,000 \times (t_{comp}$ + log(2)$t_{comm})$\\
\hline
2048 & 100 & 62,500 & 4 & $t_{comp}$ + log(4)$t_{comm}$ & $62,500 \times (t_{comp}$ + log(4)$t_{comm})$\\
\hline
4096 & 100 & 31,250 & 8 & $t_{comp}$ + log(8)$t_{comm}$ & $31,250 \times (t_{comp}$ + log(8)$t_{comm}$)\\
\hline
8192 & 100 & 15,625 & 16 & $t_{comp}$ + log(16)$t_{comm}$ & $15,625 \times (t_{comp}$ + log(16)$t_{comm})$\\
\hline
... & ... & ... & ...\\
\hline
1,280,000 & 100 & 100 & 2500 & $t_{comp}$ + log(2500)$t_{comm}$ & $100 \times (t_{comp}$ + log(2500)$t_{comm}$)\\
\hline
\end{tabular}
\label{table:large_batch_analysis}
\end{table*}

\begin{table}[ht]
\footnotesize
\renewcommand{\arraystretch}{1.3}
\caption{Standard Benchmarks for ImageNet training.}
\centering
\begin{tabular}{|c|cc|}
\hline
Model & Epochs & Test Top-1 Accuracy \\
\hline
\hline
AlexNet & 100 & 58\% \cite{iandola2016firecaffe} \\
\hline
ResNet-50 & 90 & 75.3\% \cite{he2016deep} \\
\hline
\end{tabular}
\label{table:standard_benchmark}
\end{table}

\section{Large-Batch DNN Training}

\subsection{Benefits of Large-Batch Training}
The asynchronous methods using parameter server are not guaranteed to be stable on large-scale systems \cite{chen2016revisiting}.
As discussed in~\cite{goyal2017accurate}, data-parallelism synchronized approach is more stable for very large DNN training. The idea is simple---by using a {\bf large batch size} for SGD, the work for each iteration can be easily distributed
to multiple processors.  
%The most stable method for large-scale DNN training is data-parallelism synchronous approach \cite{goyal2017accurate}.
%As mentioned before, the DNN training is very slow. 
%, it takes 14 days to finish the ResNet-50 training for ImageNet dataset by one NVIDIA M40 GPUs. 
Consider the following ideal case. ResNet-50 requires 7.72 billion single-precision operations to process one 225x225 image. If we run 90 epochs for ImageNet dataset, the number of operations is 90 * 1.28 Million * 7.72 Billion ($10^{18}$). Currently, the most powerful supercomputer can finish $200 \times 10^{15}$ single-precision operations per second \cite{dongarra2017}. If there is an algorithm allowing us to make full use of the supercomputer, we can finish the ResNet-50 training in 5 seconds.

To do so, we need to make the algorithm use more processors and load more data at each iteration, which corresponds
to increasing the batch size in synchronous SGD. 
Let us use one NVIDIA M40 GPU to illustrate the case of a single machine.
In a certain range, larger batch size will make the single GPU's speed higher (Figure \ref{fig:gpu_speed}). The reason is that low-level matrix computation libraries will be more efficient.
For ImageNet training with Alexthe Net model the, optimal batch size per GPU is 512. 
If we want to use many GPUs and make each GPU efficient, we need a larger batch size.
For example, if we have 16 GPUs, then we should set the batch size  to $16 \times 512 = 8192$.
Ideally, if we fix total number of data accesses and grow the batch size linearly with number of processors, 
the number of SGD iterations will decrease linearly and the time cost of each iteration remains constant, 
so the total time will also reduce linearly with number of processors (Table \ref{table:large_batch_analysis}). 

\subsection{Model Selection}
To scale up the algorithm to many machines, a major overhead is the communication among different machines \cite{zhang2015deep}.
Here we define {\bf scaling ratio}, which means the ratio between computation and communication.
For DNN models, the computation is proportional to the number of floating point operations required for processing an image.
Since we focus on synchronous SGD approach, the communication is proportional to model size (or the number of parameters). 
Different DNN models have different scaling ratios.
To generalize our study, we pick two representative models: AlexNet and ResNet50. The reason is that they have different scaling ratios.
From Table \ref{tab:scaling_ratio}, we observe that ResNet50's scaling ratio is 12.5$\times$ larger than that of AlexNet.
This means scaling ResNet50 is easier than scaling AlexNet.
Generally, ResNet50 will have a much higher weak scaling efficiency than AlexNet. 

In the fixed-epoch situation, large batch does not change the number of floating point operations (computation volume).
However, large batch can reduce the communication volume. The reason is that the single-iteration communication volume is only related to the model size and network system.
Larger batch size means less number of iterations and less overall communication. 
Thus, large batch size can improve the algorithm's scalability. 

\subsection{Difficulty of Large-Batch Training}
\begin{figure*}[htb!]
    \footnotesize 
    %\centering
    \begin{subfigure}[t]{0.50\textwidth}
        \centering
        \includegraphics[height=2.8in]{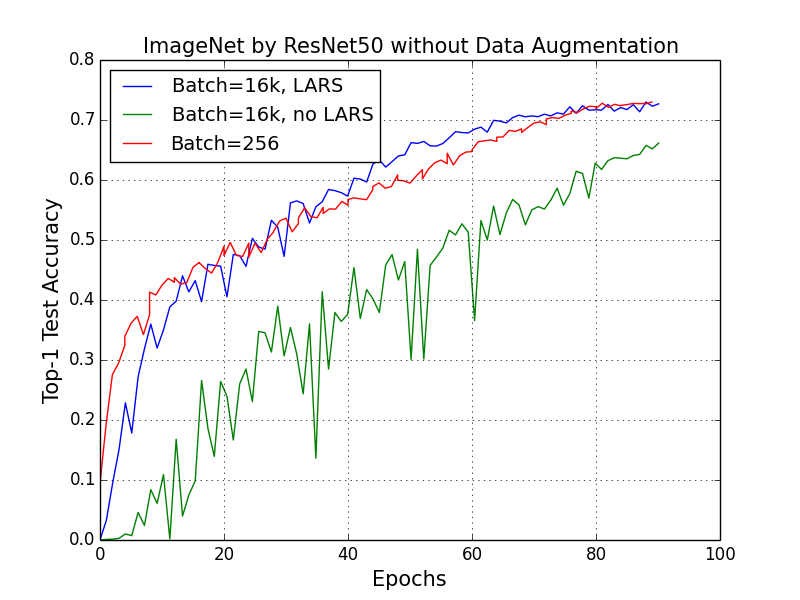}
        \caption{Batch Size=16k}
    \end{subfigure}
    \begin{subfigure}[t]{0.50\textwidth}
        \centering
        \includegraphics[height=2.8in]{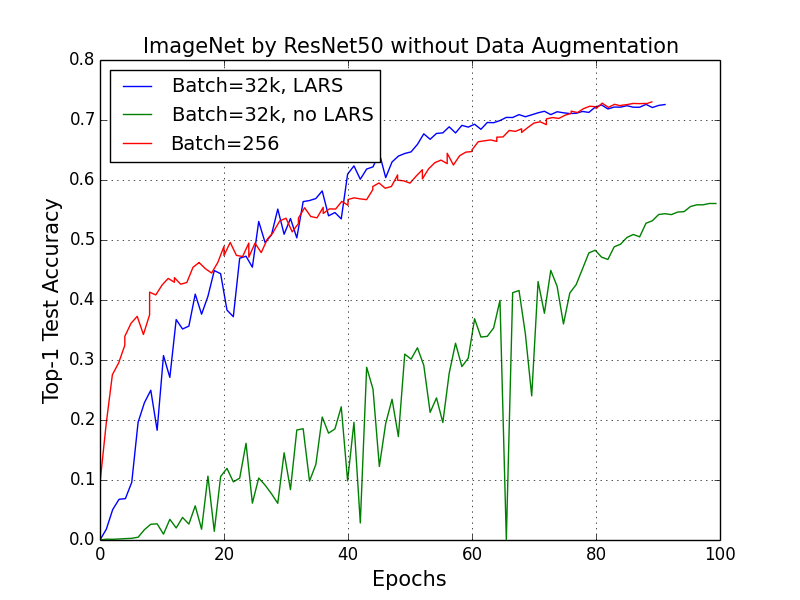}
        \caption{Batch Size=32k}
    \end{subfigure}
    \caption{\footnotesize \label{fig:resnet50_lars_effects}The base learning rate of Batch 256 is 0.2 with poly policy (power=2). For the version without LARS, we use the state-of-the-art approach \cite{goyal2017accurate}: 5-epoch warmup and linear scaling for LR. For the version with LARS, we also use 5-epoch warmup. Clearly, the existing method does not work for Batch Size larger than 8K. LARS algorithm can help the large-batch to achieve the same accuracy with baseline in the same number of epochs.}
\end{figure*}

However, synchronous SGD with larger batch size usually achieves lower accuracy than when used with smaller batch sizes, if each is run for the same
number of epochs, and currently there is no algorithm allowing us to effectively use very  large batch sizes. \cite{keskar2016large}. 
%
%there is no general algorithm that allows us to use very large batch size \cite{keskar2016large}. 
%Generally, the large-batch case's test accuracy will be much lower than that of the small-batch case (the baseline) if they run the same number of epochs.
Table \ref{table:standard_benchmark} shows the target accuracy by standard benchmarks.
%We use the ImageNet training to illustrate the difficulty of large-batch training.
For example, when we set the batch size of AlexNet larger than 1024 or the batch size of ResNet-50 larger than 8192, the test accuracy will be significantly decreased (Table \ref{table:alexnet_4k} and Figure \ref{fig:resnet50_lars_effects}).

For large-batch training, we need to ensure that the larger batches achieve similar test accuracy with the smaller batches by running the same number of epochs. 
Here we fix the number of epochs because: Statistically, one epoch means the algorithm touches the entire dataset once; and computationally, fixing the number of epochs means fixing the number of floating point operations. 
State-of-the-art approaches for large batch training include two techniques:

(1) {\bf Linear Scaling} \cite{krizhevsky2014one}: If we increase the batch size from $B$ to $kB$, we should also increase the learning rate from $\eta$ to $k\eta$.

(2) {\bf Warmup Scheme} \cite{goyal2017accurate}: If we use a large learning rate ($\eta$). We should start from a small $\eta$ and increase it to the large $\eta$ in the first few epochs.

The intuition of linear scaling is related to the number of iterations. Let us use $B$, $\eta$, and $I$ to denote the batch size, the learning rate, and the number of iterations. If we increase the the batch size from $B$ to $kB$, then the number of iterations is reduced from $I$ to $I/k$. This means that the frequency of weight updating reduced by $k$ times. Thus, we make the updating of each iteration $k\times$ more efficient by enlarging the learning rate by $k$ times. 
The purpose of a warmup scheme is to avoid the situation in which the algorithm diverges at the beginning because we have to use a very large learning rate based on linear scaling.
With these techniques, researchers can use the relatively large batch in a certain range (Table \ref{table:state_of_the_art}). However, we observe that state-of-the-art approaches can only scale batch size to 1024 for AlexNet and 8192 for ResNet-50.
If we increase the batch size to 4096 for AlexNet, we only achieve 53.1\% in 100 epochs (Table \ref{table:alexnet_4k}). Our target is to achieve 58\% accuracy even when using large batch sizes.

\begin{table*}[ht]
\footnotesize
\renewcommand{\arraystretch}{1.3}
\caption{State-of-the-art large-batch training and test accuracy.}
\centering
\begin{tabular}{|c|ccccc|}
\hline
Team & Model & Baseline Batch & Large Batch & Baseline Accuracy & Large Batch Accuracy\\
\hline
\hline
Google \cite{krizhevsky2014one} & AlexNet & 128 & 1024 & 57.7\% & 56.7\%\\
\hline
%Google\footnote{\tiny Chen et al, \textit{Revisiting Distributed Synchronous SGD}, 2016 (Google Report)} & GoogleNet & 1600 & 6400 & NO & NO\\
%\hline
Amazon \cite{li2017scaling} & ResNet-152 & 256 & 5120 & 77.8\% & 77.8\%\\
\hline
Facebook \cite{goyal2017accurate} & ResNet-50 & 256 & 8192 & 76.40\% & 76.26\%\\
\hline
\end{tabular}
\label{table:state_of_the_art}
\end{table*}

\begin{table}[ht]
\footnotesize
\renewcommand{\arraystretch}{1.3}
\caption{Current approaches (linear scaling + warmup) do not work for AlexNet with a batch size larger than 1024. We tune the warmup epochs from 0 to 10 and pick the one with highest accuracy. According to linear scaling, the optimal learning rate (LR) of batch size 4096 should be 0.16. We use poly learning rate policy, and the poly power is 2. The momentum is 0.9 and the weight decay is 0.0005.}
\centering
\begin{tabular}{|c|cccc|}
\hline
Batch Size & Base LR & warmup & epochs & test accuracy \\
\hline
\hline
512 & 0.02 & N/A & 100 & 0.583\\
\hline
1024 & 0.02 & no & 100 & 0.582\\
\hline
4096 & 0.01 & yes & 100 & 0.509\\
\hline
4096 & 0.02 & yes & 100 & 0.527\\
\hline
4096 & 0.03 & yes & 100 & 0.520\\
\hline
4096 & 0.04 & yes & 100 & 0.530\\
\hline
4096 & 0.05 & yes & 100 & 0.531\\
\hline
4096 & 0.06 & yes & 100 & 0.516\\
\hline
4096 & 0.07 & yes & 100 & {\bf 0.001}\\
\hline
... & ... & ... & ... & ...\\
\hline
4096 & 0.16 & yes & 100 & {\bf 0.001}\\
\hline
\end{tabular}
\label{table:alexnet_4k}
\end{table}

\begin{table}
 \footnotesize
  \caption{Scaling Ratio for AlexNet and ResNet50.}
  \label{tab:scaling_ratio}
\centering
    \vspace{3pt}
  \begin{tabular}{|c|c|c|c|}
  \hline
  Model & communication & computation & comp/comm\\
   & \# parameters & \# flops per image & scaling ratio\\
    \hline
    \hline
   AlexNet & \# 61 million & \# 1.5 billion & 24.6\\
    \hline
   ResNet50 & \# 25 million & \# 7.7 billion & 308\\
    \hline

  \end{tabular}
\end{table}

\begin{table}[tb]
\footnotesize
  %\vspace{-10pt}
  \caption{ImageNet Dataset with AlexNet Model. We use ploy learning rate policy, and the poly power is 2. The momentum is 0.9 and the weight decay is 0.0005. For a batch size of 32K, we changed local response norm in AlexNet to batch norm. Specifically, we use the refined AlexNet model by B. Ginsburg$^1$.}
  \label{tab:alexnet_auto_lr}
\centering
    \vspace{3pt}
  \begin{tabular}{|c|cccc|}
    \hline
    Batch Size & LR rule & warmup & Epochs & test accuracy\\
    \hline
    512 & regular & N/A & 100 & 0.583\\
    \hline
    4096 & LARS & 13 epochs & 100 & 0.584\\
    \hline
    8192 & LARS & 8 epochs & 100 & 0.583\\
    \hline
    32768 & LARS & 5 epochs & 100 & 0.585\\
    \hline
  \end{tabular}
%  \vspace{-15pt}
\end{table}

 \begin{table*}
 \footnotesize
  \caption{For batch size=32K, we changed local response norm in AlexNet to batch norm.}
  \label{tab:alexnet_speed_cost}
\centering
    \vspace{3pt}
  \begin{tabular}{|c|c|c|c|c|c|}
  \hline
  Batch Size & epochs & Peak Top-1 Accuracy & hardware & time\\
    \hline
    \hline
    256 & 100 & 58.7\% & 8-core CPU + K20 GPU & 144h\\
   \hline
   512 & 100 & 58.8\% & DGX-1 station & 6h 10m\\
   \hline
   4096 & 100 & 58.4\% & DGX-1 station & 2h 19m\\
   \hline
   32768 & 100 & 58.5\% & 512 KNLs & 24m\\
   \hline
   32768 & 100 & 58.6\% & 1024 CPUs & 11m\\
   \hline
  \end{tabular}
\end{table*}

\begin{table*}
 \footnotesize
  \caption{ResNet50 Results. We use the same data augmentation with the original ResNet-50 model \cite{he2016deep}. When we use the batch size equal to 32768, we finished the 90-epoch ImageNet training in 20 minutes without losing accuracy.}
  \label{tab:resnet50_speed_cost}
\centering
    \vspace{3pt}
  \begin{tabular}{|c|c|c|c|c|c|}
  \hline
  Batch Size & Data Augmentation & epochs & Peak Top-1 Accuracy & hardware & time\\
    \hline
    \hline
    256 & NO & 90 & 73.0\% & DGX-1 station & 21h\\
   \hline
   256 & YES & 90 & 75.3\% & 16 KNLs & 45h\\
   \hline
   8192 & NO & 90 & 72.7\% & DGX-1 station & 21h\\
   \hline
   8192 & NO & 90 & 72.7\% & 32 CPUs + 256 P100 GPUs & 1h\\
   \hline
   8192 & YES & 90 & 75.3\% & 32 CPUs + 256 P100 GPUs & 1h\\
   \hline
    16384 & YES & 90 & 75.3\% & 1024 CPUs & 52m\\
   \hline
    {\bf 16000} & {\bf YES} & {\bf 90} & {\bf 75.3\%} & {\bf 1600 CPUs} & {\bf 31m}\\
   \hline
   32768 & NO & 90 & 72.6\% & 512 KNLs & 1h\\
   \hline
    32768 & YES & 90 & 75.4\% & 512 KNLs & 1h\\
   \hline
    32768 & YES & 90 & 75.4\% & 1024 CPUs & 48m\\
   \hline
    {\bf 32768} & {\bf YES} & {\bf 90} & {\bf 75.4\%} & {\bf 2048 KNLs} & {\bf 20m}\\
   \hline
    {\bf 32768} & {\bf YES} & {\bf 64} & {\bf 74.9\%} & {\bf 2048 KNLs} & {\bf 14m}\\
   \hline
  \end{tabular}
\end{table*}

\begin{table*}
 \footnotesize
  \caption{Overall Comparison by 90-epoch ResNet50 Top-1 Val Accuracy.}
  \label{tab:resnet50_compare}
\centering
    \vspace{3pt}
  \begin{tabular}{|c|c|c|c|c|c|c|}
  \hline
  Batch Size & 256 & 8K & 16K & 32K & 64K & note\\
    \hline
    \hline
    MSRA & 75.3\% & 75.3\% &  --- & --- & --- & weak data augmentation\\
   \hline
   IBM & --- & 75.0\% & --- & --- & --- & ---\\
   \hline
   SURFsara & --- & 75.3\% & --- & --- & --- & ---\\
   \hline
   Facebook & 76.3\% & 76.2\% & 75.2\% & 72.4\% & 66.0\% & Heavy data augmentation\\
   \hline
   Our version & 73.0\% & 72.7\% & 72.7\% & 72.6\% & 70.0\% & no data augmentation\\
   \hline
   Our version & 75.3\% & 75.3\% & 75.3\% & 75.4\% & 73.2\% & weak data augmentation\\
   \hline
  \end{tabular}
\end{table*}

\subsection{Scaling up Batch Size}
In this paper, we use LARS algorithm \cite{you2017scaling} together with warmup scheme \cite{goyal2017accurate} to scale up the batch size.
Using these two approaches, synchronous SGD with a large batch size can achieve the same accuracy as the baseline (Table \ref{tab:alexnet_auto_lr}). 
To scale to larger batch sizes (e.g. 32K) for AlexNet, we need to change the local response normalization (LRN) to batch normalization (BN). We add BN after each Convolutional layer. Specifically, we use the refined AlexNet model by B. Ginsburg\footnote{https://github.com/borisgin/nvcaffe-0.16/tree/caffe-0.16/models/alexnet\_bn}.
From Figure \ref{fig:resnet50_lars_effects}, we can clearly observe the effects of LARS. LARS can help ResNet-50 to preserve the high test accuracy. The current approaches (linear scaling and warmup) has much lower accuracy
for batch size = 16k and 32k (68\% and 56\%). The target accuracy is about 73\%.

\iffalse
\begin{figure}
\includegraphics[width=3.8in]{figs/resnet50_64k.png}
\caption{\footnotesize \label{fig:resnet50_64k}Our baseline achieves 75.3\% accuracy. We can scale the batch size to 32K, which only lost 0.6\% accuracy. The accuracy loss is larger than 1\% for the batch size larger than 32K, which is not acceptable.}
\end{figure}
\fi

\section{Experimental Results}
\subsection{Experimental Settings. }
The dataset we used in this paper is ImageNet-1k \cite{deng2009imagenet}.
The dataset has 1.28 million images for training and 50,000 images for testing.
Without data augmentation, the top-1 testing accuracy of our ResNet-50 baseline is 73\% in 90 epochs.
For versions without data augmentation, we achieve state-of-the-art accuracy (73\% in 90 epochs).
With data augmentation, our accuracy is 75.3\%. 
%In the original paper, the accuracy of ResNet-50 is 75.3\% after 90 epochs \cite{he2016deep}.
%We use the same network as the original ResNet-50 paper.
%Accuracy in this paper means top-1 test accuracy. Time means training time.
For the KNL implementation, we have two versions: 

(1) We wrote our KNL code based on Caffe \cite{jia2014caffe} for single-machine processing and use MPI \cite{gropp1996high} for the communication among different machines on KNL cluster.

(2) We use Intel Caffe, which supports multi-node training by Intel MLSL (Machine Learning Scaling Library).

%We use Intel Omni path as the interconnect network for communication. 
We use the TACC Stampede 2 supercomputer as our hardware platform\footnote{portal.tacc.utexas.edu/user-guides/stampede2}.
%Both GPU and KNL are general-purpose processors, which can be used in both DNN training and inference.
All GPU-related data are measured based on B. Ginsburg's nvcaffe\footnote{https://github.com/borisgin/nvcaffe-0.16}.
The LARS algorithm is opened source by NVIDIA Caffe 0.16. We implemented the LARS algorithm based on NVIDIA Caffe 0.16.

\subsection{ImageNet training with AlexNet}
Previously, NVIDIA reported that using one DGX-1 station they were able to finish 90-epoch ImageNet training with AlexNet in 2 hours. 
However, they used half-precision or FP16, whose cost is half of the standard single-precision operation.
We run 100-epoch ImageNet training with AlexNet with standard single-precision. It takes 6 hours 9 minutes for batch size = 512 on one NVIDIA DGX-1 station.
Because of LARS algorithm \cite{you2017scaling}, we are able to have the similar accuracy using large batch sizes (Table \ref{tab:alexnet_auto_lr}). If we increase the batch size to 4096, it only needs 2 hour 10 minutes on one NVIDIA DGX-1 station.
Thus, using large batch can significantly speedup DNN training.

For the AlexNet with batch size = 32K, we scale the algorithm to 512 KNL sockets with a total of about 32K processors cores. The batch size allocated per individual KNL socket is 64, so the overall batch size is $32678$. We finish the 100-epoch training in 24 minutes.
When we use 1024 CPUs, with a batch size per CPU of 32, we finish 100-epoch AlexNet training in 11 minutes.
To the best of our knowledge, this is currently the fastest 100-epoch ImageNet training with AlexNet.
The overall comparison is in Table \ref{tab:alexnet_speed_cost}.

\subsection{ImageNet training with ResNet-50}
Facebook \cite{goyal2017accurate} finishes the 90-epoch ImageNet training with ResNet-50 in one hour on 32 CPUs and 256 NVIDIA P100 GPUs. P100 is the processor used in the NVIDIA DGX-1.
After scaling the batch size to 32K, we are able to more KNLs. 
We use 512 KNL chips and the batch size per KNL is 64.
We finish the 90-epoch training in 32 minutes on 1600 CPUs using a batch size of 32K.
We finish the 90-epoch training in 31 minutes on 1600 CPUs using a batch size of 16,000.
We finish the 90-epoch training in 20 minutes on 2048 CPUs using a batch size of 32K.
The version of our CPU chip is Intel Xeon Platinum 8160 (Skylake).
The version of our KNL chip is Intel Xeon Phi Processor 7250.
%, which costs 2,436 USD\footnote{ark.intel.com/products/codename/48999/Knights-Landing}.
Note that we are not affiliated to Intel or NVIDIA, and we do not have any \textit{ a priori} preference for GPUs or KNL.
%We just want to show that we can achieve the same results in a 3.4 times lower budget. 
The overall comparison is in Table \ref{tab:resnet50_speed_cost}.
%We are not affiliated to Intel or NVIDIA. We do not have preference on Intel or NVIDIA. This is an open science study.
%We just want to show that we can achieve the same results in a much lower budget.

Codreanu \textit {et al.} reported their experience on using Intel KNL clusters to speed up ImageNet training by a blogpost\footnote{https://blog.surf.nl/en/imagenet-1k-training-on-intel-xeon-phi-in-less-than-40-minutes/}.
They reported that they achieved 73.78\% accuracy (with data augmentation) in less than 40 minutes on 512 KNLs.
Their batch size is 8k.
However, Codreanu \textit{et al.} only finished 37 epochs. If they conduct 90-epoch training, the time is 80 minutes with 75.25\% accuracy.
In terms of absolute speed (images per second or flops per second), Facebook and our version are much faster than Codreanu \textit{et al.}
Since both Facebook and Codreanu used data augmentation, Facebook's 90-epoch accuracy is higher than that of Codreanu.

\subsection{ResNet-50 with Data Augmentation}
Based on the original ResNet50 model \cite{he2016deep}, we added data augmentation to our baseline. 
%The top-1 val accuracy of original ResNet50 is 75.3\% in 90 epochs. 
Our baseline achieves 75.3\% top-1 val accuracy in 90 epochs. 
Because we do not have Facebook's model file, we failed to reproduce full match their results of 76.24\% top-1 accuracy.
The model we used is available upon request.
Codreanu \textit{et al.} reported they achieved 75.81\% top-1 accuracy in 90 epochs; however, they changed the model parameters (not only hyper-parameters).
The overall comparison is in Table \ref{tab:resnet50_compare}.
We observe that our scaling efficiency is much higher than Facebook's version.
Even though our baseline's accuracy is lower than Facebook's, we achieve a correspondingly higher accuracy when we increase the batch size above 10K.
%The accuracy-epoch curve of our version is shown in Figure \ref{fig:resnet50_64k}.
Akiba \textit{et al.}~\cite{akiba2017extremely} reported finishing the 90-epoch ResNet-50 training within 15 minutes on 1,024 Nvidia P100 GPUs. 
However, the baseline accuracy is missing in the report, so it is difficult to tell how much their 74.9\% accuracy using the 32k batch size diverges from the baseline. 
Secondly, both Akiba et al. and Facebook's work~\cite{goyal2017accurate} are ResNet-50 specific, while we also show the generality of our approach with AlexNet.
%which is harder to scale out due to the ratio between computation and communication (Section \ref{sec:comp_comm_ratio}). 
%Also, it is hard to keep the accuracy of AlexNet when people increase the batch size (Section~\ref{sec:difficulty}).
It is worth noting that our online preprint is two months earlier than Akiba \textit{et al.}

\subsection{NVIDIA P100 GPU and Intel KNL}
Because state-of-the-art models like ResNet50 are computational intensive, our comparison is focused on the computational power rather than memory efficiency.
Since deep learning applications mainly use single-precision operations, we do not consider double-precision here.
The peak performance of P100 GPU is 10.6 Tflops\footnote{http://www.nvidia.com/object/tesla-p100.html}.
The peak performance of Intel KNL is 6 Tflops\footnote{https://www.alcf.anl.gov/files/HC27.25.710-Knights-Landing-Sodani-Intel.pdf}.
Based on our experience, the power of one P100 GPU is roughly equal to two KNLs. 
For example, we used 512 KNLs to match Facebook's 256 P100 GPUs.
However, using more KNLs still requires the larger batch size.

\subsection{Scaling Efficiency of Large Batches}
To scale up deep learning, we need to a communication-efficient approach.
Communication means moving data. On a shared memory system, communication means moving data between different level of memories (e.g. from DRAM to cache). 
On a distributed system, communication means moving the data over the network (e.g. master machine broadcast its data to all the worker machines).
Communication often is the major overhead when we scale the algorithm on many processors. Communication is much slower than computation (Table \ref{tab:alphabeta}). Also, communication costs much more energy than computation (Table \ref{tab:energy_cost}). 

\begin{figure}[!t]
\centering
\includegraphics[width=3.0in]{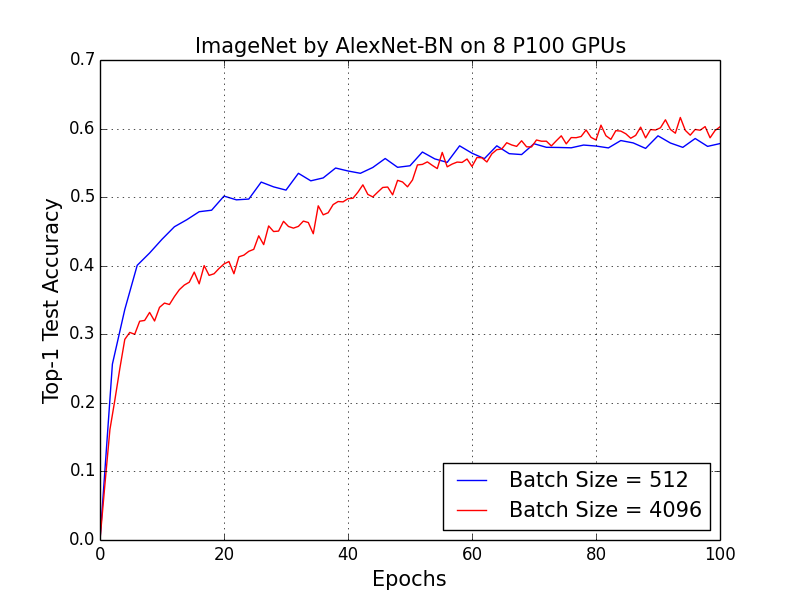}
\caption{From this figure, we observe that we can achieve the target accuracy in the same number of epochs by using large batch size. Batch Size = 512 is the baseline.}
\label{fig:alexnet_epoch}
\end{figure}

\begin{figure}[!t]
\centering
\includegraphics[width=3.0in]{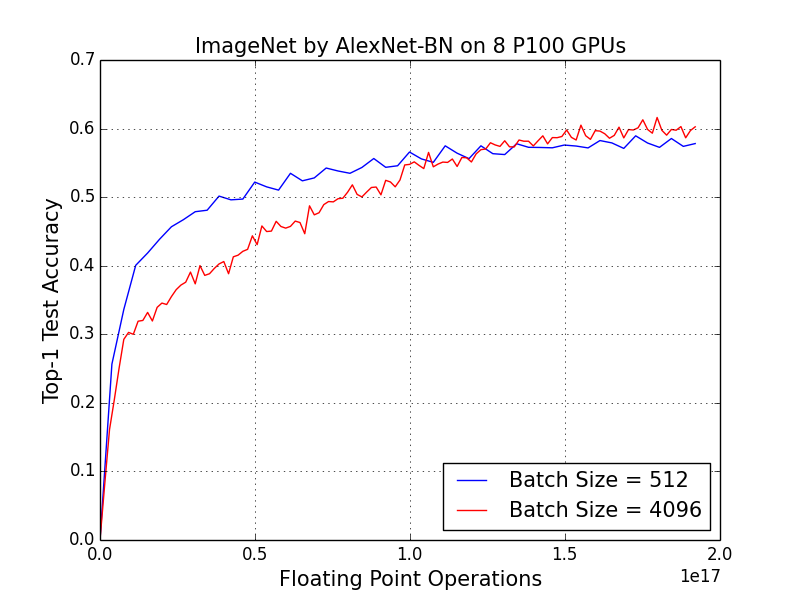}
\caption{Increasing the batch size does not increase the number of floating point operations. Large batch can achieve the same accuracy in the fixed number of floating point operations.}
\label{fig:alexnet_flops}
\end{figure}

\begin{figure}[!t]
\centering
\includegraphics[width=3.0in]{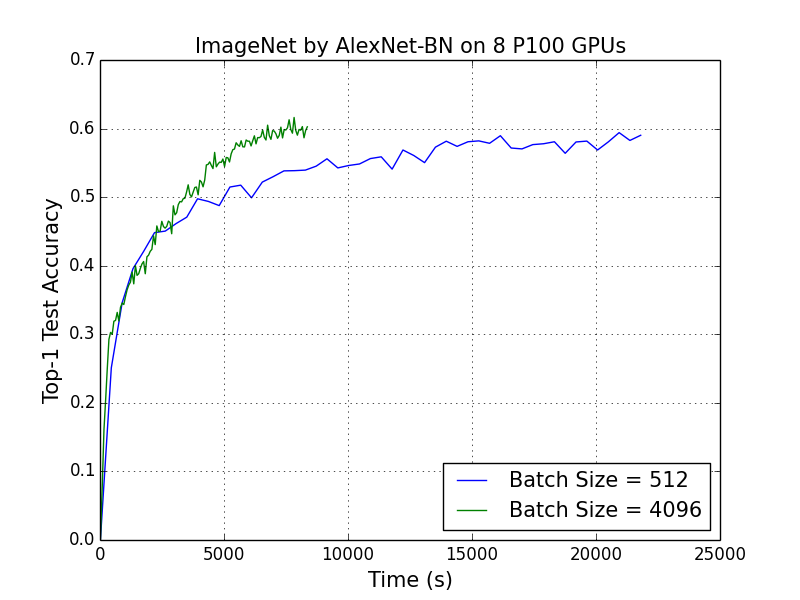}
\caption{When we have enough computational powers, the larger batch version is much faster than the smaller batch. To achieve 58\% accuracy, the larger batch (batch size = 4096) only needs about two hours while the  smaller batch (batch size = 512) needs about six hours. The large batch and small batch versions finish the same number of floating point operations.}
\label{fig:alexnet_time}
\end{figure}

\begin{figure}[!t]
\centering
\includegraphics[width=3.0in]{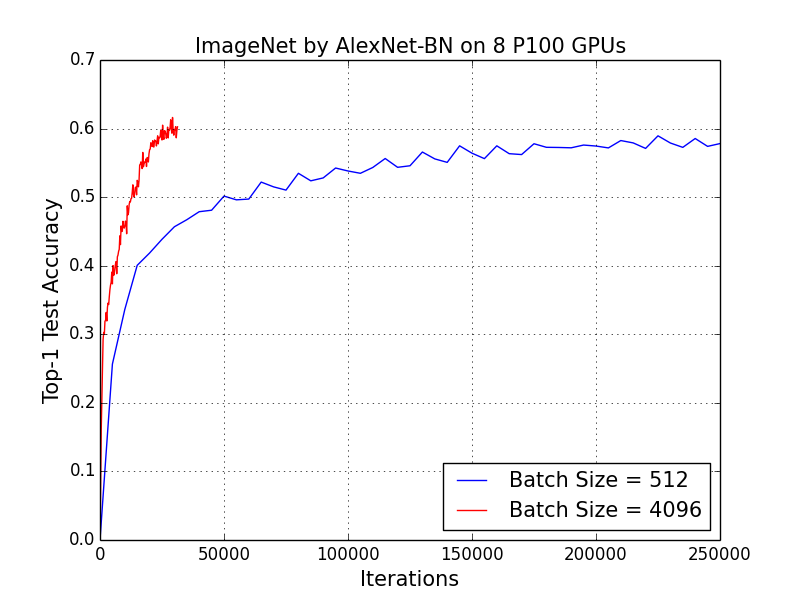}
\caption{When we fix the number of epochs and increase the batch size, we need much less iterations.}
\label{fig:alexnet_iter}
\end{figure}

\begin{figure}[!t]
\centering
\includegraphics[width=3.0in]{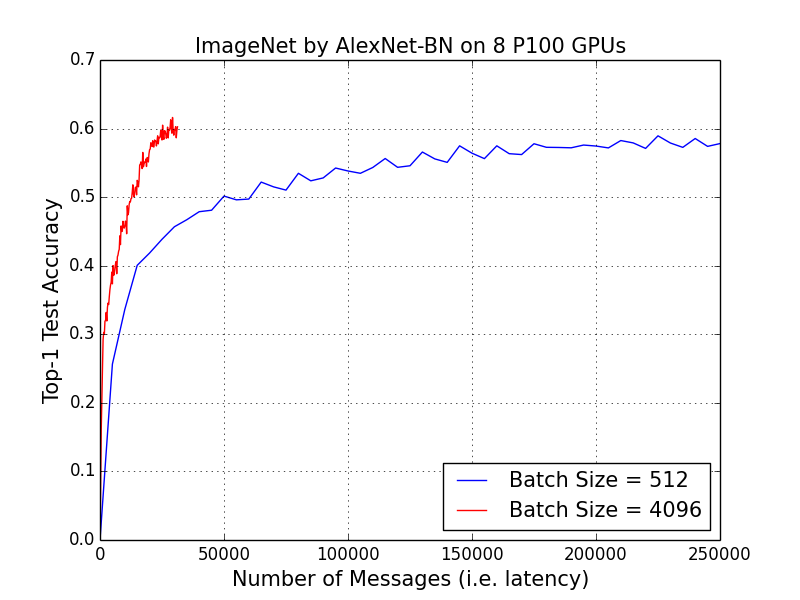}
\caption{When we fix the number of epochs and increase the batch size, we need much less iterations. The number of iterations is linear with the number of messages the algorithm sent.}
\label{fig:alexnet_message}
\end{figure}

\begin{figure}[!t]
\centering
\includegraphics[width=3.0in]{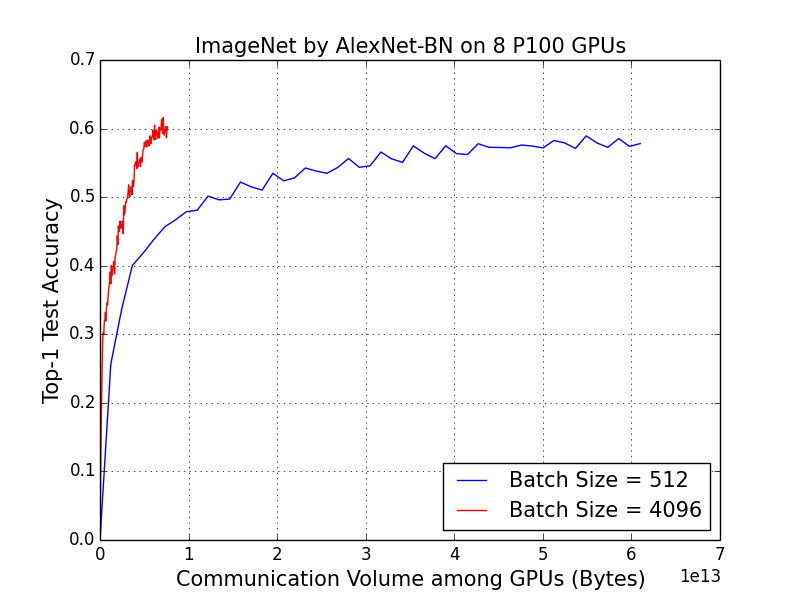}
\caption{Let us use $E$, $n$, and $B$ to denote the number of epochs, the total number of pictures in the training dataset, and the batch size, respectively. Then we can get the number of iterations is $E \times n / B$. When we fix the number of epochs and increase the batch size, we need much less iterations. The number of iterations is linear with the number of messages the algorithm sent. Let us denote $|W|$ as the neural network model size. Then we can get the communication volume is $|W| \times E \times n / B$. Thus, the larger batch version needs to move much less data than the smaller batch when they finish the number of floating point operations.}
\label{fig:alexnet_data}
\end{figure}

Let us use the example ImageNet training with AlexNet-BN on 8 P100 GPUs to illustrate the idea. The baseline's batch size is 512. The larger batch size is 4096.
In this example, we focus the the communication among different GPUs.
Firstly, our target is to make training with the larger batch size achieve the same accuracy as the small batch in the same fixed number of epochs 
(Figure \ref{fig:alexnet_epoch}). Fixing the number of epochs implies fixing the number of floating point operations (Figure \ref{fig:alexnet_flops}). 
If the system is not overloaded, the larger batch implementation is much faster than small batch for using the same hardware (Figure \ref{fig:alexnet_time}).
For finishing the same number of epochs, the communication overhead is lower in the large batch version than in the smaller batch version. Specifically, the larger batch version sends fewer messages (latency overhead) and moves less data (bandwidth overhead) than the small batch version. 
For Sync SGD, the algorithm needs to conduct an all-reduce operations (sum of gradients on all machines). The number of messages sent is linear with the number of iterations. Also, because the gradients has the same size with the weights ($|W|$). 
Let us use $E$, $n$, and $B$ to denote the number of epochs, the total number of pictures in the training dataset, and the batch size, respectively. Then we can get the number of iterations is $E \times n / B$. Thus, when we increase the batch size, we need much less number of iterations (Table \ref{table:large_batch_analysis} and Figure \ref{fig:alexnet_iter}). The number of iterations is equal to the number of messages the algorithm sent (i.e. latency overhead). Let us denote $|W|$ as the neural network model size. Then we can get the communication volume is $|W| \times E \times n / B$. Thus, the larger batch version needs to move much less data than smaller batch version when they finish the number of floating point operations (Figures \ref{fig:alexnet_message} and \ref{fig:alexnet_data}).
In summary, the larger batch size does not change the number of floating point operations when we fix the number of epochs. The larger batch size can increase the computation-communication ratio because it reduces the communication overhead (reduce latency and move less data).
Finally, the larger batch size makes the algorithm more scalable on distributed systems.

\begin{table}
  \footnotesize
  \caption{Communication is much slower than computation because time-per-flop ($\gamma$)  $\ll$  1/ bandwidth ($\beta$)  $\ll$  latency ($\alpha$). For example, $\gamma = 0.9 \times 10^{-13}$s for NVIDIA P100 GPUs.}
  \label{tab:alphabeta}
  \begin{tabular}{|c|c|c|}
  \hline
    Network & $\alpha$ (latency) & $\beta$ (1/bandwidth)\\
    \hline
    \hline
    Mellanox 56Gb/s FDR IB & $0.7\times10^{-6}$s & $0.2\times10^{-9}$s\\
    \hline
    Intel 40Gb/s QDR IB & $1.2\times10^{-6}$s & $0.3\times10^{-9}$s\\
    \hline
    Intel 10GbE NetEffect NE020 & $7.2\times10^{-6}$s & $0.9\times10^{-9}$s\\
      \hline
\end{tabular}
\end{table} 

\begin{table}
  \footnotesize
  \centering
  \caption{Energy table for 45nm CMOS process \cite{horowitzenergy}. Communication costs much more energy than computation.}
  \label{tab:energy_cost}
  \begin{tabular}{|c|c|c|}
  \hline
    Operation & Type & Energy (pJ)\\
    \hline
    \hline
    32 bit int add & Computation & 0.1\\
    \hline
    32 bit float add & Computation & 0.9\\
    \hline
    32 bit register access & Communication & 1.0\\
    \hline
    32 bit int multiply & Computation & 3.1\\
    \hline
    32 bit float multiply & Computation & 3.7\\
    \hline
    32 bit SRAM access & Communication & 5.0\\
    \hline
    32 bit DRAM access & Communication & 640\\
    \hline
\end{tabular}
\end{table} 

\section{Conclusion}

In recent years the ImageNet 1K benchmark set has played a significant role as a benchmark for assessing different approaches to DNN training. The most successful results on accelerating DNN training on ImageNet have used a synchronous SGD approach. 
To scale this synchronous SGD approach to more processors requires increasing the batch size. 
Using a warm-up scheme coupled with a linear scaling rule, researchers at Facebook \cite{goyal2017accurate} were able to scale the training of ResNet 50 to 256 Nvidia P100's with a batch size of 8K and a total training time of one hour. 
Using a more sophisticated approach to adapting the learning rate in a method they named  the Layer-wise Adaptive Rate Scaling (LARS) algorithm \cite{you2017scaling}, researchers were able to scale the batch size to 32K; however, the potential for scaling to larger number of processors was not demonstrated in that work, and only 8 Nvidia P100 GPUs were employed. Also, data augmentation was not used in that work, and accuracy was impacted.
In this paper we confirmed that the increased batch sizes afforded by the LARS algorithm could lead to increased scaling. 
In particular, we scaled synchronous SGD batch size to 32K and using 1024 Intel Skylake CPUs we were able to finish the 100-epoch ImageNet training with AlexNet in 11 minutes. 
Furthermore, with a batch size of 32K and 2048 KNLs we were able to finish 90-epoch ImageNet training with ResNet-50 in 20 minutes. 
State-of-the-art ImageNet training speed with ResNet-50 is 74.9\% top-1 test accuracy in 15 minutes \cite{akiba2017extremely}.
We got 74.9\% top-1 test accuracy in 64 epochs, which only needs {\bf 14 minutes}.
%If we use the batch size of 16,000 and 1600 CPUs, we are able to finish 90-epoch ImageNet training with ResNet-50 in 31 minutes without losing accuracy. 
We also explored the impact of data augmentation in our work. 
%In comparison our platform of 512 Intel %KNLs, Facebook's platform includes 32 CPUs %and 256 NVIDIA P100 GPUs.

\section{Acknowledgements}
The large batch training algorithm was developed jointly with I.Gitman and B.Ginsburg done during Yang You's internship at NVIDIA in the summer 2017.
The work presented in this paper  was supported by the National Science Foundation, through the Stampede 2 (OAC-1540931) award.
JD and YY are supported by the U.S. DOE Office of Science, Office of Advanced Scientific Computing Research, Applied Mathematics program under Award Number DE-SC0010200; by the U.S. DOE Office of Science, Office of Advanced Scientific Computing Research under Award Numbers DE-SC0008700; by DARPA Award Number HR0011-12- 2-0016, ASPIRE Lab industrial sponsors and affiliates Intel, Google, HP, Huawei, LGE, Nokia, NVIDIA, Oracle and Samsung. Other industrial sponsors include Mathworks and Cray. In addition to ASPIRE sponsors, KK is supported by an auxiliary Deep Learning ISRA from  Intel. 
CJH also thank XSEDE and Nvidia for independent support.

%\clearpage

\bibliographystyle{aaai}
\bibliography{ref}

\end{document}